\documentclass[letterpaper, 10 pt, conference]{ieeeconf}  

\IEEEoverridecommandlockouts                              

\usepackage{cite}
\usepackage[colorlinks=true,allcolors=blue]{hyperref}

\usepackage{amsmath,amssymb,amsfonts}
\usepackage{mathtools}
\usepackage{bm}

\usepackage{pifont}
\newcommand{\cmark}{\ding{51}}%
\newcommand{\xmark}{\ding{55}}%

\usepackage{algorithm}
\usepackage{algorithmic}

\usepackage{array}
\usepackage{wasysym}

\usepackage{graphicx}
\usepackage[thinc]{esdiff} 
\usepackage{textcomp}
\usepackage{xcolor}
\usepackage{multirow}
\usepackage{etoolbox}
\usepackage{siunitx}
\usepackage{nth}

\usepackage[font=small]{caption}
\usepackage{url}            
\usepackage{booktabs}       
\usepackage{tabularx}

\usepackage{subfigure}

\def\BibTeX{{\rm B\kern-.05em{\sc i\kern-.025em b}\kern-.08em
    T\kern-.1667em\lower.7ex\hbox{E}\kern-.125emX}}
\newcommand{\presub}[2]{\prescript{}{#1}{#2}}

\renewcommand{\baselinestretch}{.829} 

\usepackage{acro}
\newcommand{\newac}[2]{\DeclareAcronym{#1}{short=#1,long=#2}}
\newac{CC}{Constant Curvature}
\newac{CS}{Constant Strain}
\newac{COM}{Center of Mass}
\newac{CON}{Coupled Oscillator Network}
\newac{DCM}{Discretized Cosserat Rod Model}
\newac{DOF}{Degrees of Freedom}
\newac{EOM}{Equations of Motion}
\newac{FEM}{Finite Element Method}
\newac{GVS}{Geometric Variable Strain}
\newac{LNN}{Lagrangian Neural Network}
\newac{LSTM}{Long Short-Term Memory}
\newac{MAE}{Mean Absolute Error}
\newac{ML}{Machine Learning}
\newac{MSE}{Mean Squared Error}
\newac{MPC}{Model Predictive Control}
\newac{NODE}{Neural ODE}
\newac{PAC}{Piecewise Affine Curvature}
\newac{PCS}{Piecewise Constant Strain}
\newac{PCC}{Piecewise Constant Curvature}
\newac{PDE}{Partial Differential Equation}
\newac{RL}{Reinforcement Learning}
\newac{RNN}{Recurrent Neural Network}
\newac{RMSE}{Root Mean-Squared Error}

\title{\LARGE \bf
    Learning Low-Dimensional Strain Models of Soft Robots by Looking at the Evolution of Their Shape with Application to Model-Based Control
}

\author{
Ricardo Valadas$^{*, 1}$, Maximilian Stölzle$^{*, 1, 2}$, Jingyue Liu$^{1}$, Cosimo Della Santina$^{1}$ 
\thanks{$^*$The R. Valadas and M. Stölzle contributed equally. M. Stölzle is the corresponding author.}
\thanks{The work was supported under the European Union's Horizon Europe Program from Project EMERGE - Grant Agreement No. 101070918. The contribution by M. Stölzle was additionally supported by the Dutch Cultuurfonds Wetenschapsbeurzen 2024.}
\thanks{$^{1}$All authors are with the Cognitive Robotics department, Delft University of Technology, Delft, Netherlands {\tt\scriptsize \{M.W.Stolzle, J.Liu-14, C.DellaSantina\}@tudelft.nl}. $^{2}$M. Stölzle is also affiliated with the Laboratory for Information \& Decision Systems, Massachusetts Institute of Technology, Cambridge, USA {\tt\scriptsize mstolzle@mit.edu}.}
}

\begin{document}

\bstctlcite{IEEEexample:BSTcontrol}

\maketitle
\thispagestyle{empty}
\pagestyle{empty}

\begin{abstract}
Obtaining dynamic models of continuum soft robots is central to the analysis and control of soft robots, and researchers have devoted much attention to the challenge of proposing both data-driven and first-principle solutions. Both avenues have, however, shown their limitations; the former lacks structure and performs poorly outside training data, while the latter requires significant simplifications and extensive expert knowledge to be used in practice. This paper introduces a streamlined method for learning low-dimensional, physics-based models that are both accurate and easy to interpret. We start with an algorithm that uses image data (i.e., shape evolutions) to determine the minimal necessary segments for describing a soft robot's movement. Following this, we apply a dynamic regression and strain sparsification algorithm to identify relevant strains and define the model's dynamics. We validate our approach through simulations with various planar soft manipulators, comparing its performance against other learning strategies, showing that our models are both computationally efficient and 25x more accurate on out-of-training distribution inputs. Finally, we demonstrate that thanks to the capability of the method of generating physically compatible models, the learned models can be straightforwardly combined with model-based control policies.
\end{abstract}

\section{Introduction}
Continuum soft robot's inherent compliance and embodied intelligence make them promising candidates for close collaboration between humans and robots and contact-rich manipulation~\cite{rus2015design, mengaldo2022concise}.
Modeling their dynamical behavior~\cite{armanini2023soft} with computationally tractable models is important for many applications, such as efficient simulation~\cite{alkayas2025soft}, model-based control~\cite{della2023model}, state estimation~\cite{shao2023model}, and co-design~\cite{wang2024diffusebot}. 


Developing such (low-dimensional) dynamic models is challenging and is an active area of research~\cite{alora2023data, armanini2023soft}. The use of data-driven approaches has been extensively investigated in this context~\cite{thuruthel2017learning, bruder2020data, alora2023data, chen2024data}.
These learned models exhibit poor extrapolation performance~\cite{kim2021review}, a lack of interpretability and (physical) structure preventing us from directly leveraging closed-form control solutions such as the PD+feedforward~\cite{della2023model}. Instead, researchers had to fall back to computationally expensive planning methods such as \ac{MPC}~\cite{bruder2020data, alora2023data}.


\begin{figure}[ht]
    \center
    \centerline{\includegraphics[width=1.0\columnwidth]{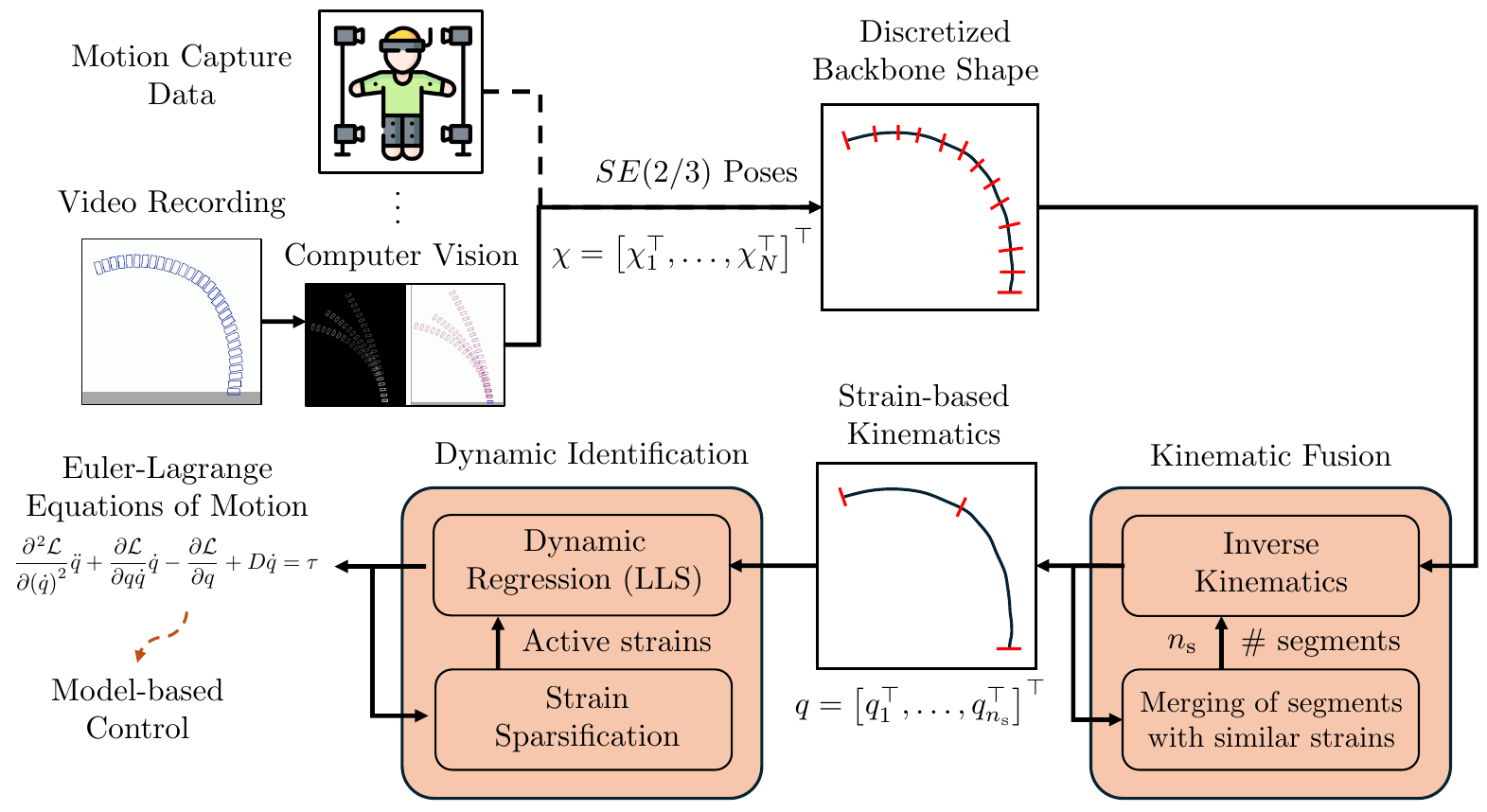}}
    \caption{
    Overview of the proposed methodology with the key contributions (\emph{Kinematic Fusion} and \emph{Dynamic Regression and Strain Sparsification}) highlighted in orange. 
    \textbf{Inputs:} We consider $N$ Cartesian pose measurements $\chi$ distributed along the soft robot backbone, for example, obtained using Computer Vision (CV) techniques from video recordings as inputs.
    \textbf{Kinematic Fusion:} We apply an iterative procedure that involves (i) computing the robot configuration $q$ using \ac{PCS} inverse kinematics, and (ii), to avoid overly complex and high-dimensional models, we merge adjacent segments with similar strains across the dataset into one segment of constant strain.
    \textbf{Dynamic Identification:} We identify the PCS dynamic model by iteratively regressing coefficients using linear least squares and further reduce the model complexity by neglecting insignificant strains.
    \textbf{Output:} The identified dynamic model has a Lagrangian structure suitable, for example, for model-based control applications.
    }
    \label{fig:overall_diag}
    \vspace{-0.2cm}
\end{figure}

The traditional avenue established by the robotics and continuum dynamics communities has been to derive the dynamical model directly from first principles~\cite{renda2018discrete, boyer2020dynamics, della2023model, armanini2023soft} 
which provides physical interpretability and structure at the cost of needing substantial expert knowledge, for example in the selection of the proper kinematic approximations (e.g., \ac{PCC}~\cite{webster2010design}, \ac{PCS}~\cite{renda2018discrete}, \ac{GVS}~\cite{boyer2020dynamics}). 
%
Suboptimal choices or even errors in applying this modeling procedure can lead to significant issues like inaccurate predictions and overly complex models. This hinders the democratization of soft robots, as only specialized research labs possess the required expertise~\cite{aracri2024soft}.

Very recently, there has been a community push towards integrating physical structures and stability guarantees into learned models (e.g., Lagrangian Neural Networks~\cite{liu2024physics}, residual dynamical formulations~\cite{bruder2024koopman, gao2024sim}, or oscillatory networks~\cite{stolzle2024input}) which combine benefits from both worlds: they are learned directly from data which reduces the expert knowledge that is needed but at the same time exhibit a physical structure that can be exploited for model-based control and stability analysis.
This work positions itself in this new trend of research, specifically focusing on deriving kinematic and dynamic models for continuum soft robots in a data-driven way.
%
%

%
Indeed, deriving reduced-order kinematic representations remains the core challenge in physics-based modeling. 
Previous works have relied heavily on the modeling engineer's intuition and experience to make decisions on the number of \ac{PCS} segments, the length of each segment, and which strains to consider~\cite{toshimitsu2021sopra}. However, these decisions are not straightforward and could easily result in models that are higher-dimensional than necessary, or that important strains are ignored based on a wrong intuition~\cite{garg2022kinematic}.
Very recently, Alkayas et al.~\cite{alkayas2025soft} proposed a data-driven algorithm to identify the optimal discrete \ac{GVS}~\cite{boyer2020dynamics} strain basis of continuum soft robots via Proper Orthogonal Decomposition (POD). 
However, the discrete strain basis requires careful numerical integration at runtime, and identifying the dynamical parameters (e.g., stiffness, damping coefficients) of the soft robot relies upon solving a nonlinear least-squares problem that is not always well behaved~\cite{stolzle2023experimental}.

This paper proposes to solve these challenges by introducing an end-to-end approach that can automatically learn both a \ac{PCS} kinematic parametrization and the corresponding dynamical model, including its dynamic parameters directly from image/Cartesian pose data.
First, a kinematic fusion algorithm aims to minimize the \acp{DOF} of the \ac{PCS} kinematic model while preserving a desired shape reconstruction accuracy for the given discrete shape measurements in Cartesian space. In contrast to previous work~\cite{alkayas2025soft}, we do not necessitate prior knowledge about strain discontinuities.
Secondly, an integrated strategy is proposed to simultaneously sparsify the strains of the \ac{PCS} model and estimate the parameters of the dynamical model with closed-form linear least-squares.
Contrary to common symbolic regression approaches such as SINDy~\cite{kaiser2018sparse}, we crucially preserve the (physical) structure of the Euler-Lagrangian dynamics as derived according to the \ac{PCS} model.
This feature allows the derived dynamical model to be subsequently rapidly deployed within established model-based controllers~\cite{della2023model}.

We verify the approach in simulation in a diverse set of scenarios, including different robot topologies and the performance when measurement noise is present. Impressively, the method is able to accurately perform long-horizon (\SI{7}{s}) shape predictions when being trained on \SI{4}{s} of trajectory data.
We benchmark the proposed approach against several state-of-the-art dynamical model learning approaches (e.g., \acp{NODE}, \ac{CON}, \acp{LNN}). On the training set, our proposed method exhibits a \SI{70}{\percent} lower shape prediction error than the best-performing baseline method (\ac{NODE}).
However, we find that the difference is even greater in extrapolation scenarios (i.e., actuation sequences and magnitudes unseen during training). Here, our proposed method reduces the shape prediction error on the test set by \SI{96}{\percent} compared to the best performing \ac{ML} baseline (\ac{NODE}).
Finally, we demonstrate how the Lagrangian structure of the identified dynamical model allows us to easily design a model-based controller that is effective at regulating the shape of the soft robot.

A video attachment presents the research idea \& methodology and contains supplementary plots and animations of the results presented in the paper.\footnote{{\small \url{https://youtu.be/dfO-PhDIiHI}}}

\section{Preliminaries}
In the following, we will provide some background on the \ac{PCS} kinematic model and the derivation of soft robot dynamics following a Lagrangian approach, which are two fundamental topics for this research paper.

\subsection{Piecewise Constant Strain (PCS) Kinematics}\label{sec:pcs_model}
The \ac{PCS} model~\cite{renda2018discrete} describes the kinematics of continuum soft robots by assuming that the six elemental local backbone strains (shear, axial, bending, twist) are piecewise constant across $n_\mathrm{s}$ segments but variable in time.
We remark that other popular kinematic models for soft robots, such as Piecewise Constant Curvature (PCC)~\cite{webster2010design}, are often times a special case of the \ac{PCS} model.


For the planar case, the configuration of the $i$th segment is referred to as $q_i= \begin{bmatrix}
    \kappa_{\mathrm{be},i} & \sigma_{\mathrm{sh},i} & \sigma_{\mathrm{ax},i}
\end{bmatrix}^\top \in \mathbb{R}^3$, where $\kappa_{\mathrm{be},i}, \sigma_{\mathrm{sh},i}, \sigma_{\mathrm{ax},i}$ are the bending, shear, and axial strains, respectively, and $i \in \{1, \dots, n_\mathrm{s} \}$. 
Therefore, the configuration of the entire soft robot is defined as $q \in \mathbb{R}^{n_\mathrm{q}}$, where $n_\mathrm{q} = 3 n_\mathrm{s}$.
We also have access to closed-form expressions for the forward and inverse kinematics of a single constant strain segment~\cite{stolzle2023experimental}.
As a consequence, forward and inverse kinematics for the entire planar \ac{PCS} soft robot can be implemented using an iterative procedure starting at the proximal end without having to resort to differential (inverse) kinematic techniques.
Namely, the forward kinematics $\pi: \mathbb{R}^{n_\mathrm{q}} \to SE(2)$ allow us to compute the pose $\chi_j = \begin{bmatrix}
    p_{\mathrm{x},j} & p_{\mathrm{y},j} & \theta_j
\end{bmatrix}^\top = \vartheta(q,s_j)$, where $s_j \in [0, L]$ is the backbone abscissa/coordinate, $L$ is the length of the entire continuum structure in an undeformed configuration, $p_{\mathrm{x},j}, p_{\mathrm{y},j} \in \mathbb{R}$ and $\theta_j$ are the positions and orientations at point $s$, respectively.
Given $N$ poses along the backbone, we can also define the inverse kinematic mapping $\varrho: 3N \times N \to n_\mathrm{q}$ that provides us with the configuration $q = \varrho(\chi, s)$ in closed-form. Here, $q_i$ will describe the configuration of the $i$th constant strain segment connecting the $i-1$th and the $i$th markers with associated poses $\chi_{i-1}$ and $\chi_i$. This means that $n_\mathrm{s} = N -1$ in order that $\rho$ can be bijective.

\subsection{Lagrangian Dynamics}\label{sub:lagr_dynamics}
The Lagrangian of a mechanical system as a function of the configuration $q \in \mathbb{R}^{n_\mathrm{q}}$ and the corresponding time derivative $\dot{q} \in \mathbb{R}^{n_\mathrm{q}}$ can be expressed as
\begin{equation}\small
    \mathcal{L}(q, \dot{q}) = \mathcal{T}(q, \dot{q}) - \mathcal{U}(q) = \underbrace{\frac{1}{2}\dot{q}^\top \, M(q) \, \dot{q}}_{\mathcal{T}(q, \dot{q})} - \underbrace{\frac{1}{2} q^\top K q - \int G(q) \: \mathrm{d}q}_{\mathcal{U}(q)},
\end{equation}
where $\mathcal{T}(q, \dot{q})$, $\mathcal{U}(q)$ are the kinetic and potential energy of the system, respectively, and $M(q) \succ 0 \in \mathbb{R}^{n_\mathrm{q} \times n_\mathrm{q}}$ is referred to as the mass matrix. $G(q) \in \mathbb{R}^{n_\mathrm{q}}$ contribute the gravitational forces and $K \in \mathbb{R}^{n_\mathrm{q} \times n_\mathrm{q}}$ represents the linear elastic stiffness of the system.
Subsequently, the Euler-Lagrangian equation can be leveraged to derive the \ac{EOM} of continuum soft robots as~\cite{della2023model, liu2024physics}
\begin{equation}\small
    \diffp[2]{\mathcal{L}}{{\dot{q}}}\ddot{q} + \diffp{\mathcal{L}}{{q}{\dot{q}}}\dot{q} - \diffp{\mathcal{L}}{{q}} + D\dot{q} = \tau
\end{equation}
where we are also considering the generalized dissipative forces $D\dot{q}$ and the actuation torques $\tau \in \mathbb{R}^{n_\mathrm{q}}$. 
In this work, we assume, without loss of generality, that both the stiffness matrix $K = \mathrm{diag}(k_1, \dots, k_{n_\mathrm{q}})$ and the damping matrix $D = \mathrm{diag}(d_1, \dots, d_{n_\mathrm{q}}) \succeq 0$ are diagonal.

\begin{figure*}[ht]
    \center
    \subfigure[Kinematic Fusion Scheme]{\includegraphics[width=0.34\textwidth]{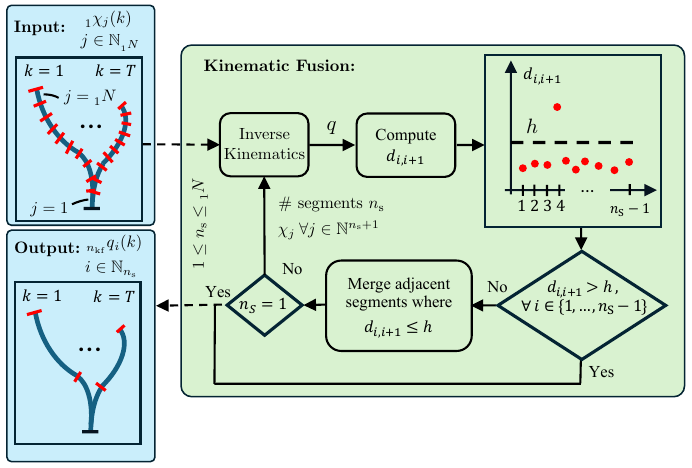}\label{fig:kin_regr}}
    \hfill
    \subfigure[Dynamic Regression \& Strain Sparsification Scheme]{\includegraphics[width=0.65\textwidth]{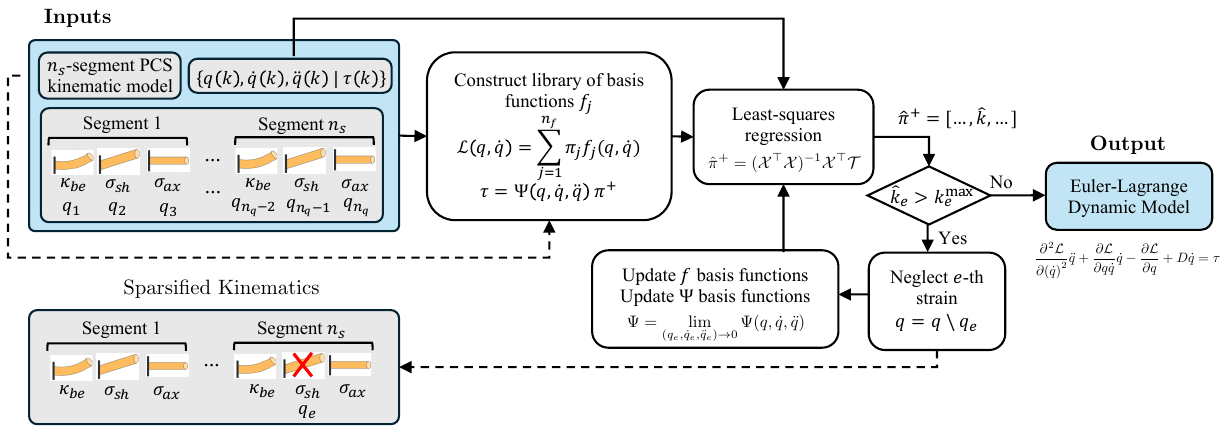}\label{fig:dyn_regr}}
    \vspace{-0.2cm}
    \caption{\textbf{Panel (a):} Schematic of the kinematic fusion algorithm. As inputs serve a sequence of $N$ discrete poses along the backbone of the soft robot. Next, we execute inverse kinematics in closed form with a $n_\mathrm{s} = N-1$ segment \ac{PCS} model to identify the (unmerged) configuration of the robot. Initially, each constant strain segment connects two neighboring backbone poses. Subsequently, we compute a strain similarity measure $\bar{d}_{i,i+1}$ between each pair of adjacent segments. If segments exhibit a similar strain (i.e., the metric falls below a threshold $h$), we merge them into one constant strain segment. This process is repeated until no more merging is possible, resulting in a kinematic model with (hopefully) fewer segments: $1 \leq n_\mathrm{s} \leq N-1$. 
    \textbf{Panel (b):} Schematic of the dynamic model identification process that simultaneously regresses the dynamic parameters and neglects unimportant strains. Based on a $n_\mathrm{s}$-segment PCS model, a library of basis functions is constructed to parameterize the system’s Lagrangian and \ac{EOM}. A regression framework is established on a dataset of configuration-space positions $\dot{q}(k)$, velocities $\dot{q}(k)$, accelerations $\ddot{q}(k)$, and actuation torques $\tau(k)$ that estimates the dynamic parameters $\hat{\pi}^+$ with closed-form, linear least squares. Strains that exhibit a stiffness higher than a predefined threshold are neglected, prompting adjustments to the basis functions. Subsequently, this procedure is repeated until all strain stiffnesses lie below the threshold.}
    \vspace{-0.5cm}
\end{figure*}
\section{Methodology}
In this work, we propose a strategy for automatically identifying low-dimensional strain models for continuum robots directly from shape trajectories, as outlined in Figure \ref{fig:overall_diag}.
We assume that we have access to the poses of $N$ markers along the backbone that represent a discretized shape description of the soft robot.
In this work, we primarily focus on the planar case, where we extract SE(2) poses using computer vision techniques. However, the proposed \emph{Kinematic Fusion} approach, along with the \emph{Dynamic Regression and Strain Sparsification} strategy, can be extended to 3D scenarios with SE(3) inputs. While obtaining consistent and unoccluded SE(3) poses solely from vision-based information in 3D can be challenging, it is feasible~\cite{zheng2024vision}. Additionally, SE(3) pose measurements can always be acquired through other proprioceptive~\cite{rosi2022sensing} or exteroceptive methods, such as motion capture systems.

The goal is now to identify kinematic and dynamical models that allow (a) to represent the shape with 
$n_\mathrm{s}$ \ac{PCS} segments~\cite{renda2018discrete}, where necessarily the final $n_\mathrm{s} \ll N$, and (b) to predict the future shape evolution of the soft robot.
We tackle this task by (i) identifying a low-dimensional parametrization (e.g., number of constant strain segments, the length of each segment, etc.) of the kinematics over a series of static snapshots and (ii) identifying the parameters of the Lagrangian model and simultaneously eliminating strains from the model that do not have a significant effect on the shape evolution.
We refer to component (i) as the \emph{Kinematic Fusion} algorithm as it is an iterative approach to merge parts of the backbone that exhibit a similar strain into constant strain segments.
The component (ii), named \emph{Dynamic Regression \& Strain Sparsification} algorithm, is an iterative procedure that, at each iteration, first regresses in closed-form the coefficients of the dynamic using linear least-squares and then eliminates strains from the dynamic model if the stiffness associated with a strain exceeds a given threshold. The intuition here is that strain would oscillate at very high frequencies, which are usually not relevant for practical control, and that it would take very high forces to introduce a significant deflection in the strain.
The output of our approach is low-dimensional kinematic and dynamical models that preserve the physical \& \ac{PCS} strain model structures.


\subsection{Kinematic Fusion}
As previously introduced, the algorithm is provided for each training dataset sample $k \in \{1, \dots, T\}$ with $N$ pose measurements $\chi$ and associated backbone abscissas $s \in \mathbb{R}^N$ distributed along the backbone of the soft robot.
Therefore, we initialize at $l=1$: $\presub{1}{N} = N$ and $\presub{1}{\chi} = \chi$, where $l \in \mathbb{N}_{\geq 1}$ denotes the iteration index.

At the beginning of each iteration, we leverage the closed-form inverse kinematics to compute the configuration $\presub{l}{q} = \rho(\presub{l}{\chi}, \presub{l}{s}) $ of a $\presub{l}{n}_\mathrm{s} = \presub{l}{N} - 1$ segment \ac{PCS} model. We now compute between each of the total $\presub{l}{n}_\mathrm{s} - 1$ segment pairs, the following normalized strain similarity measure
\begin{equation}\small
    \presub{l}{\bar{d}}_{i,i+1} = \frac{1}{T} \sum_{k=1}^T \left \lVert \frac{\presub{l}{q}_{i+1}(k) - \presub{l}{q}_i(k)}{q^\mathrm{max} - q^\mathrm{min}} \right \rVert_2, \quad \forall i \in \{1, \presub{l}{n}_\mathrm{s}-1 \},
\end{equation}
where
\begin{equation}\small
    q^\mathrm{min} = \min_{i \in \mathbb{N}_{n}, k \in \mathbb{N}_T} q_i(k), 
    \qquad 
    q^\mathrm{max} = \max_{i \in \mathbb{N}_{n}, k \in \mathbb{N}_T} q_i(k),
\end{equation}
are the minimum and maximum configuration values across the dataset, respectively. This normalization is necessary as strains usually exhibit vastly different magnitudes. For example, the bending strains are usually more than one order of magnitude larger than the axial strains.
We keep all segment pairs with $\presub{l}{\bar{d}}_{i,i+1} > h$, where $h$ is a tunable threshold, separate. Oppositely, we merge all neighboring/adjacent segments with $\presub{l}{\bar{d}}_{i,i+1} \leq h$ into one single segment of constant strain.
If indeed $\exists \: i \in \{ \mathbb{N}_{\presub{l}{n}_\mathrm{s}} | \presub{l}{\bar{d}}_{i,i+1} < h \}$, then then kinematic model is reduced to $\presub{l+1}{n}_\mathrm{s}$ segments, where $\presub{l+1}{n}_\mathrm{s} < \presub{l}{n}_\mathrm{s}$.
As a final step of the iteration, we now subsample the Cartesian poses $\presub{l+1}{\chi}$ and the associate backbone coordinates $\presub{l+1}{s}$ such that they only contain the tip of each of the fused segments.

The kinematic fusion step is repeated with $l = l + 1$ for $n_\mathrm{kf}-1$ iterations until no more merging is possible, which can occur either if $\presub{n_\mathrm{kf}}{{\bar{d}}}_{i,i+1} > h \: \forall i \in \{1, \presub{n_\mathrm{kf}}{n}_\mathrm{s}-1 \}$ (i.e., the strain similarity measure is larger than $h$ for every pair of segments) or the model gets reduced to a single segment (i.e., $\presub{n_\mathrm{kf}}{n}_\mathrm{s} = 1$).
We illustrate the kinematic fusion algorithm in Fig.~\ref{fig:kin_regr}, and an example of the thresholding is visualized in Fig.~\ref{fig:results:kinematic_fusion:pcs_ns-2}.

\subsection{Dynamic Identification: Dynamic Regression \& Strain Sparsification}
After obtaining a kinematic model with the \emph{Kinematic Fusion} algorithm, we can employ the identified parametrization as a foundation for deriving a dynamic model. 
First, we symbolically derive the basis functions of the \ac{PCS} dynamical model.
Subsequently, we implement an iterative procedure to (i) regress dynamic coefficients with linear least squares and (ii) identify strains that can be neglected and remove them from the dynamical model.

\subsubsection{Parametrization of the PCS Dynamic Model with Basis Functions}
In order to easily regress the dynamic parameters with linear least squares, we first derive the \ac{PCS} dynamics for a $n_\mathrm{s}$ segment soft robot from first principle~\cite{armanini2023soft, della2023model}, with all rotational and linear strains taken into account, and subsequently parametrize both the Lagrangian and Euler-Lagrangian equations as a linear combination of monomial basis function.
\begin{equation}\label{eq:lagrangian_basis_functions}\small
    \mathcal{L}(q, \dot{q}) = \sum_{j=1}^{n_\mathrm{f}} \pi_j f_j(q, \dot{q}), 
    \quad
    \tau = \sum_{j=1}^{n_\psi} \pi_j^{+}\Psi_j(q,\dot{q},\ddot{q}) \in \mathbb{R}^{n_q},
\end{equation}
where $f_j: \mathbb{R}^{n_\mathrm{q}} \times \mathbb{R}^{n_\mathrm{q}} \to \mathbb{R}$ denotes each of the basis functions, and $\pi \in \mathbb{R}^{n_\mathrm{f}}$ the corresponding coefficients.
Analog, we derive symbolically the \ac{EOM} using an Euler-Lagrangian approach (see Section~\ref{sub:lagr_dynamics}) and now state the corresponding basis functions as $\psi(q, \dot{q}, \ddot{q}) \in \mathbb{R}^{n_\Psi \times n_\mathrm{q}}$ with $\psi_j: \mathbb{R}^{n_\mathrm{q}} \times \mathbb{R}^{n_\mathrm{q}} \times \mathbb{R}^{n_\mathrm{q}} \to \mathbb{R}^{n_\mathrm{q}}$ such that
\begin{equation}\small
    \tau = \sum_{j=1}^{n_{f}} \left[ \pi_j \left( \diffp[2]{f_j}{{\dot{q}}}\ddot{q} + \diffp{f_j}{{q}{\dot{q}}}\dot{q} - \diffp{f_j}{{q}} \right) \right] + D\dot{q},
\end{equation}
where
$\pi^+ = \begin{bmatrix}
    \pi^\top & d^\top 
\end{bmatrix}^\top \in \mathbb{R}^{n_{\psi}}$ contains the associated coefficients and consists of $\pi$ and the damping coefficients $d = \mathrm{diag}(D) \in \mathbb{R}^{n_\mathrm{q}}$.

\subsubsection{Regression of Dynamic Parameters}\label{ssub:reg_dyn_params}

In order to estimate the dynamic coefficients, we formulate the linear regression problem as $\mathcal{T}=X \pi^+$, which accommodates the dataset of positions, velocities, and accelerations 
$\mathcal{X} = [\Psi(q(1), \dot{q}(1), \ddot{q}(1))^{\top},...,\Psi(q(T), \dot{q}(T), \ddot{q}(T))^{\top}]^{\top} \in \mathbb{R}^{T n_\mathrm{q} \times n_{\psi}}$ 
and the corresponding actuation inputs 
$\mathcal{T}=[\tau^{\top}(1), \dots,\tau^{\top}(T)] \in \mathbb{R}^{T n_\mathrm{q}}$. 
We solve this optimization problem with linear least squares, which minimizes the residual error as $\min \lVert \mathcal{T} - \mathcal{X} \, \pi^+ \rVert_2^2$ and allows us to identify the dynamic model coefficients in closed form as
\begin{equation}
    \hat{\pi}^+ = (\mathcal{X}^{\top} \, \mathcal{X})^{-1} \, \mathcal{X}^{\top} \mathcal{T}.
\end{equation}

\subsubsection{Sparsification of Strains}\label{ssub:strain_spars}
This dynamic identification method offers the advantage of having interpretable results, as each estimated coefficient has some physical meaning within the \ac{PCS} dynamic derivation. Specifically, among those we can extract the estimated stiffness matrix $\hat{K} = \mathrm{diag}(\hat{k}_1, \dots, \hat{k}_{n_\mathrm{q}})$, allowing us to assess the importance of each strain through its stiffness magnitude $\hat{k}_e$. A strain with high stiffness usually exhibits low displacement, approximating rigid behavior. Therefore, such strain can be considered non-essential and neglected in the dynamics.
We define a maximum allowable stiffness $k^\mathrm{max}_e \in \mathbb{R}_{\geq0}$ for each strain/configuration variable as a function of the maximum Elastic and Shear moduli $E^{\text{max}}, G^{\text{max}} \in \mathbb{R}_{\geq 0}$. For example, in the planar case and for constant cross sections of area $A_\mathrm{c}$ and second moment of inertia $I_\mathrm{c}$, this can be conveniently done as
\begin{equation}\small
    k_{\mathrm{be}}^{\mathrm{max}} = I_c E^{\mathrm{max}},
    \quad
    k_{\text{sh}}^{\text{max}} = A_c G^{\mathrm{max}},
    \quad
    k_{\text{ax}}^{\text{max}} = A_c E^{\text{max}}.
\end{equation}

Given these maximum stiffnesses, the $e$-th strain is neglected if its estimated stiffness lies above the threshold $\hat{k}_e>k_e^{\mathrm{max}}$. Therefore, the $e$-th strain is removed as a configuration variable $q = q\setminus q_e$, and its influence on the dynamics needs to be eliminated as well. We update the Euler-Lagrange basis functions as $\Psi = \lim_{(q_e,\dot{q}_e,\ddot{q}_e)\to 0} \Psi (q, \dot{q}, \ddot{q})$. Any columns that turn into all-\emph{zeros} are also removed, and the coefficient vector $\pi^+$ is updated accordingly by removing the corresponding rows. A similar procedure applies to the Lagrangian basis functions $f$ and their coefficients $\pi$.


This procedure of regressing dynamic coefficients and sparsifying strains, as presented in Sections~\ref{ssub:reg_dyn_params} and \ref{ssub:strain_spars}, respectively, is the number of strains/configuration variables converges (i.e., remains constant). We stress that the (likely) computationally expensive operation of symbolically deriving the library of basis function only needs to be once at the beginning as we subsequently update the library by taking the limit at the end of each iteration.


\section{Validation}
To validate the proposed approach, we verify the kinematic and dynamic regression algorithms separately.
We test the kinematic fusion algorithm on simulated continuum soft robots that behave according to the \ac{PCS} and \ac{PAC} model.
Subsequently, we compare the dynamic prediction performance of the proposed method against multiple \ac{ML} baseline methods on various \ac{PCS} soft robots.
Finally, we demonstrate how the regressed methods can be used in a plug-and-play fashion within a closed-form, model-based setpoint regulation framework.

\subsection{Experimental Setup}

\subsubsection{Evaluation Cases}
\textbf{Case 1: 1S PCS}, \textbf{Case 2: 2S PCS} and \textbf{Case 3: 3S PCS} represent one-, two- and three-segment planar \ac{PCS} soft robots ($n_\mathrm{s} \in \{ 1, 2, 3 \}$), respectively, with configurations $q \in \mathbb{R}^{n_\mathrm{q}}$ where $n_\mathrm{q} \in \{3, 6, 9 \}$ and actuation  $\tau \in \mathbb{R}^3, \mathbb{R}^6, \mathbb{R}^9$, assuming full actuation.
\textbf{Case 4: 1S PCS H-SH} is a one-segment \ac{PCS} robot with high shear stiffness three orders of magnitude larger than in \emph{Case 1}.
\textbf{Case 5: 2S PCS H-AX/SH} is a two-segment \ac{PCS} robot, where the \nth{1} segment has a significantly increased axial stiffness and \nth{2} segment an increased shear stiffness w.r.t \emph{Case 2}.
\textbf{Case 6: 1S PAC} considers a one-segment \ac{PAC} robot whose curvature can be described by an affine function~\cite{stella2023piecewise}.

\subsubsection{Dataset Generation}
In order to illustrate the end-to-end nature of our proposed method, we generate the datasets as short video sequences of the soft robot's movement. 
Therefore, we mimic a camera placed parallel to the robot's plane of motion to capture the soft robot's ground-truth dynamics.
At each time step, we render an image of the soft robot that contains $N=21$ equally distant, visually salient features. In the real world, this could be achieved by attaching markers to the soft robot that allows tracking of points along the backbone across time~\cite{stella2022experimental}.
We simulate the robot's ground-truth dynamics using the planar \ac{PCS} simulator presented in \cite{stolzle2023experimental}.

For Cases 1 to 4, we include eight trajectories with randomly sampled initial conditions in the training set. We consider stepwise actuation sequences for which we randomly sample a torque every \SI{10}{ms}. Each trajectory produces a \SI{0.5}{s} video captured at \SI{1000}{Hz}. For \emph{Case 6}, since the PAC simulator only accounts for kinematics, we generate an image sequence featuring the robot in $500$ randomly selected configurations.
As the test set, an additional trajectory with \SI{7}{s} duration is generated by applying a sinusoidal actuation sequence with $\tau = a_1\sin{(\omega_1 t)} + a_2\cos{(\omega_2 t)} \in \mathbb{R}^{n_\mathrm{q}}$, where $a_1$ and $a_2$ are random amplitudes, and $\omega_1$ and $\omega_2$ are random frequencies.

\subsubsection{Backbone Shape Detection from Images}
To apply our proposed model identification method, we first extract the motion of several Cartesian-space samples along the robot's backbone.
As we consider a planar problem setting and rendered images of the soft robot's shape, the goal is to extract the SE(2) poses of $N$ cross-sections along the robot.
To satisfy the assumption behind the \emph{Kinematic Fusion} algorithm, the number of extracted poses $N$ should be significantly larger than the expected number of \ac{PCS} segments required to model the robot's behavior accurately: $N \gg n_\mathrm{s}$.
%
We leverage the \emph{OpenCV} library for detecting the soft robot contour (\texttt{findContours}) and extracting pose measurements along its backbone (\texttt{minAreaRect}).
Each frame is binarized, and the contours of cross-sections are identified.
This allows the extraction of the center position $(p_{\mathrm{x},j},p_{\mathrm{y},j})$ and orientation $\theta_j$ of each cross-section (also referred to as \emph{marker}).
As in our case, the markers are equally distant, we compute, without loss of generality, the backbone abscissa as $s_j= \frac{j-1}{N} L$. 
For $T$ video frames, this results in a time sequence of SE(2) poses $\{\chi (1),...,\chi (T) \}$,  $\chi_j = \begin{bmatrix}
    \chi_1^\top \dots \chi_N^\top
\end{bmatrix}^\top \in \mathbb{R}^{3N}$, and $j \in \{ 1, \dots, N \}$.
We leverage a Savitzky-Golay filter (\nth{3}-order, window length $25$) to estimate the associated pose velocities $\dot{\chi}$ and velocities $\ddot{\chi}$.


\subsubsection{Evaluation metrics}
To evaluate the models quantitatively, we introduce position and orientation task-space metrics. 
We use a Cartesian-space \ac{MAE} measuring the deviation of the estimated from the actual robot body shape, given by

\begin{equation}\footnotesize
\begin{split}
     e_\mathrm{p}^{\mathrm{body}} = \sum_{k=1}^T \sum_{j=1}^N \frac{\lVert \hat{p}_j(k) - p_j(k) \Vert_2}{N T},
     \:
    e_{\theta}^{\mathrm{body}} = \sum_{k=1}^T \sum_{j=1}^N \frac{| \hat{\theta}_j(k) - \theta_j(k) |}{N T},
\end{split}
\end{equation}
where 
$\hat{p}_i(k)$ and $\hat{\theta}_i(k) \in \mathbb{R}$ are the estimated position and orientation of point $i$ along the structure, respectively, while $p_i(k)$ and $\theta_i(k)$ are the ground-truth counterparts. These metrics give the average pose error across all $T$ frames of a trajectory and all $N$ cross-sections tracked along the robot, enabling a good evaluation of the kinematic model by capturing how well it represents the overall shape of the soft robot structure.

\begin{figure*}[ht]
    \centering
    \subfigure[2S PCS: Strain distances]{\includegraphics[width=0.23\textwidth,trim={5 5 5 5}]{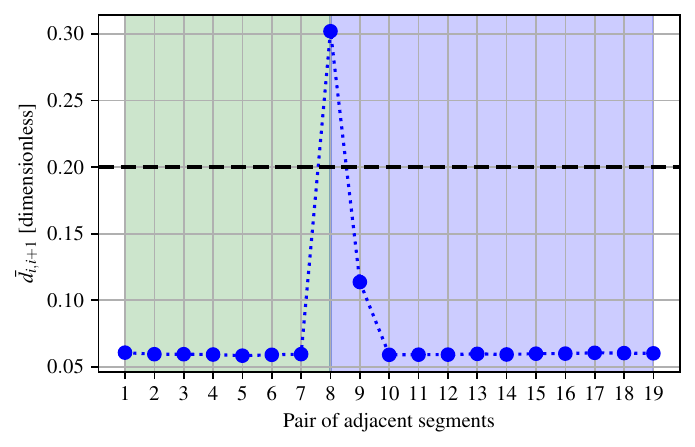}\label{fig:results:kinematic_fusion:pcs_ns-2}}
    \subfigure[3S PCS: Strain distances]{\includegraphics[width=0.23\textwidth,trim={5 5 5 5}]{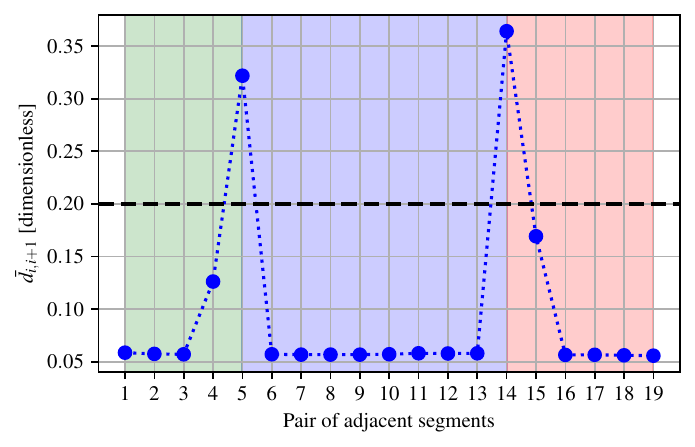}\label{fig:results:kinematic_fusion:pcs_ns-3}}
    \subfigure[1S PAC: Strain distances]{\includegraphics[width=0.23\textwidth,trim={5 5 5 5}]{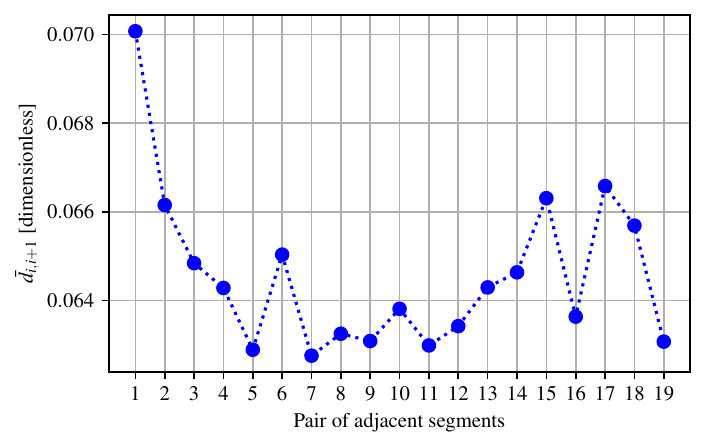}\label{fig:distances_pac}}
    \subfigure[1S PAC: Kinematic Pareto Front]{\includegraphics[width=0.25\textwidth,trim={5 5 5 5}]{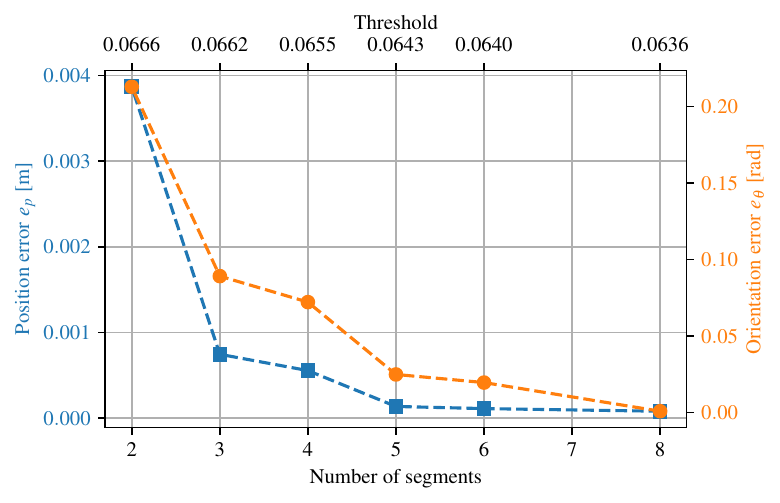}\label{fig:pareto_front_pac}}
    \vspace{-0.2cm}
    \caption{
    Kinematic Fusion Results:
    \textbf{Panels (a) \& (b):} Average strain distances between pairs of adjacent segments for \emph{Cases 2 \& 3}. The poses of $20$ markers along the manipulators are tracked, resulting in $19$ pairs of segments to be evaluated for strain similarity. The threshold is represented by a dashed line, and the background shading marks the resulting segments (separate segments are shaded in different colors).
    \textbf{Panel (c):} Average strain distances between segments after the first iteration of the kinematic fusion algorithm for \emph{Case 6}. \textbf{Panel (b):} Pareto front that describes the trade-off between the DOF of the kinematic model (i.e., the number of segments) and the shape reconstruction error for \emph{Case 6}.
    The blue and orange lines represent the average kinematic body position error $e_\mathrm{p}^\mathrm{body}$ and the body orientation error $e_\theta^\mathrm{body}$, respectively.
    The strain distance threshold $h$ that is used for separating segments is plotted on the upper x-axis.
    }\label{fig:results:kinematic_fusion}
    \vspace{-0.4cm}
\end{figure*}

In addition, we consider an end-effector Cartesian-space \ac{MAE} given by
\begin{equation}\small
    e_p^{\mathrm{ee}} = \frac{1}{ T} \sum_{k=1}^T \lVert \hat{p}_\mathrm{ee}(k) - p_\mathrm{ee}(k) \rVert_2,
    \
    e_{\theta}^{\mathrm{ee}}=\frac{1}{T} \sum_{k=1}^T | \hat{\theta}_\mathrm{ee}(k) - \theta_\mathrm{ee}(k) |,
\end{equation}
where $\hat{p}_\mathrm{ee}(k) = \hat{p}_{N}(k) \in \mathbb{R}^2$ and $\hat{\theta}_\mathrm{ee}(k) \in \mathbb{R}$ are the estimated end-effector position and orientation, respectively, with $p_\mathrm{ee}(k)$ and $\theta_\mathrm{ee}(k)$ being the ground-truth counterparts. 
This metric is particularly useful for assessing the obtained dynamic models with a control perspective, as the accuracy of the end-effector predictions is crucial for effective control in task space.

\subsection{Kinematic Fusion Results}
\subsubsection{Cases 1, 2 and 3}
Figs. \ref{fig:results:kinematic_fusion:pcs_ns-2} \& \ref{fig:results:kinematic_fusion:pcs_ns-3} presents the average strain distances between the adjacent segment pairs for \emph{Cases 2-3}, as the result of the first and final iteration of the kinematic fusion algorithm. 
The plots for Cases 2 and 3 reveal one and two peaks, respectively, revealing where we need to separate segments in the kinematic model.
The strain distance threshold $h$ is a hyperparameter that trade-offs model complexity with model accuracy.
For common soft robots that exhibit a \ac{PCS}-like behavior, we recommend choosing $h$ such as that the segmented model exhibits isolated peaks in the strain distance metric, as visible in Fig.~\ref{fig:results:kinematic_fusion}.
The number of segments and respective lengths for the resulting models obtained with a threshold of $h=0.2$ are presented in the third column of Table \ref{tab:results_kin_reg_pcs} and by comparing to the second column, it is easily visible that our algorithm almost perfectly identified the segment lengths. 
%
As a consequence of the correct identification of the number of segments and segment lengths, the identified kinematic model also accurately captures the shape of the robot with the position errors below \SI{0.2}{\percent} of the robots' length, as it can be seen from the body shape reconstruction errors stated in the first three rows of Table \ref{tab:results_kin_reg_pcs}.
We remark that the error can be further reduced by using a finer discretization of backbone pose markers (i.e., increasing $N$).

\begin{table}[htbp]
\centering
\caption{Kinematic fusion results: The second and third columns contain the actual and estimated segment lengths, respectively. The Cartesian pose error between the actual and estimated backbone shape is stated in the third and fourth columns, respectively. \emph{Case 6} represents a one-segment \ac{PAC} soft robot of total length \SI{150}{mm} and can be, therefore, not be represented with a (one-segment) \ac{PCS} model.}
\label{tab:results_kin_reg_pcs}
\setlength\tabcolsep{1.5pt}
\begin{scriptsize}
    \begin{tabular}{ccccc}\toprule
    \textbf{Case} & \begin{tabular}[c]{@{}c@{}}\textbf{Actual segment}\\ \textbf{lengths} $\mathbf{L}$ \textbf{[mm]}\end{tabular} & \begin{tabular}[c]{@{}c@{}}\textbf{Estimated segment}\\ \textbf{lengths} $\mathbf{\hat{L}}$ \textbf{[mm]}\end{tabular} & $\mathbf{e_\mathrm{p}^{\mathrm{body}}}$ \textbf{[mm]} & $\mathbf{e_{\theta}^{\mathrm{body}}}$ \textbf{[rad]} \\ \midrule
    \textbf{1: 1S PCS} & $[100]$ & $[100]$ & $0.082$ & $6.38\times 10^{-3}$ \\
    \textbf{2: 2S PCS} & $[70, 100]$ & $[68, 102]$ & $0.240$ & $1.15\times 10^{-2}$               \\
    \textbf{3: 3S PCS} & $[50, 100, 60]$ & $[52.5, 94.5, 63.0]$ & $0.210$ & $9.67\times 10^{-3}$ \\
    \textbf{6: 1S PAC} & $[150]$ & $[7.5, 105.0, 37.5]$ & $0.746$ & $8.92\times 10^{-2}$  \\
    \bottomrule
    \end{tabular}
\end{scriptsize}
\end{table}


\subsubsection{Case 6}
Even though the kinematics of many continuum soft robotic manipulators can be described by \ac{PCS}/\ac{PCC} kinematics, other continuum soft robots can only be described by piecewise constant models in the limit $N \to \infty$ as they exhibit polynomial curvature~\cite{della2019control, stella2022experimental} or even more generally \ac{GVS}~\cite{boyer2020dynamics}.
This is particularly the case when external or gravitational forces dominate the elastic and actuation forces~\cite{della2023model}.
In order to verify that our approach is also able to identify effective models in such situations, we test the kinematic fusion algorithm on the case of an affine curvature robot~\cite{stella2023piecewise} and plot the resulting average strain distances in Fig.~\ref{fig:distances_pac}.
Indeed, the strain distance plot no longer exhibits clear, isolated peaks (i.e., a single solution). Therefore, we formulate a Pareto front in Fig.~\ref{fig:pareto_front_pac} (by varying the strain distance threshold $h$) that describes the tradeoff between the number of \ac{PCS} segments (i.e., the \ac{DOF} of the kinematic model) and the shape reconstruction accuracy. Analyzing and exploiting this tradeoff allows the user to choose their \emph{sweetspot} between model complexity and performance.
In this case, we find that three segments represent a suitable compromise between model complexity and shape reconstruction accuracy as it exhibits a position error of only \SI{0.5}{\percent} of the robot's length.

\subsection{Dynamic Model Identification Results}

\subsubsection{Verification of Dynamical Regression}
We verify the dynamical regression algorithm on one and two-segment \ac{PCS} robots (i.e., Cases 1 \& 2) both with and without measurement noise.
After regressing the dynamic parameters on the training set, we perform a rollout on the test set and compare the resulting predicted trajectory with the ground truth.
To confirm that the dynamic regression is also effective when applied to real-world data, we apply in some experiments Gaussian noise that mirrors measurement noise as we would encounter it for motion capture data, computer vision detection errors, etc., to the poses included in the training set (i.e., $\Tilde{\chi} = \chi + \mathcal{N}(0, \sigma_\mathrm{n}))$.
For \emph{Case 1}, we sample the noise from a normal distribution with standard deviations \SI{0.5}{mm} and \SI{1}{\degree} for the position and orientation measurements, respectively.
Analog, we define the standard deviation of the noise for \emph{Case 2} as \SI{0.1}{mm} and \SI{0.5}{\degree}.

\begin{figure}[htbp]
    \centering
    \includegraphics[width=0.8\columnwidth, trim={5 5 5 5}]{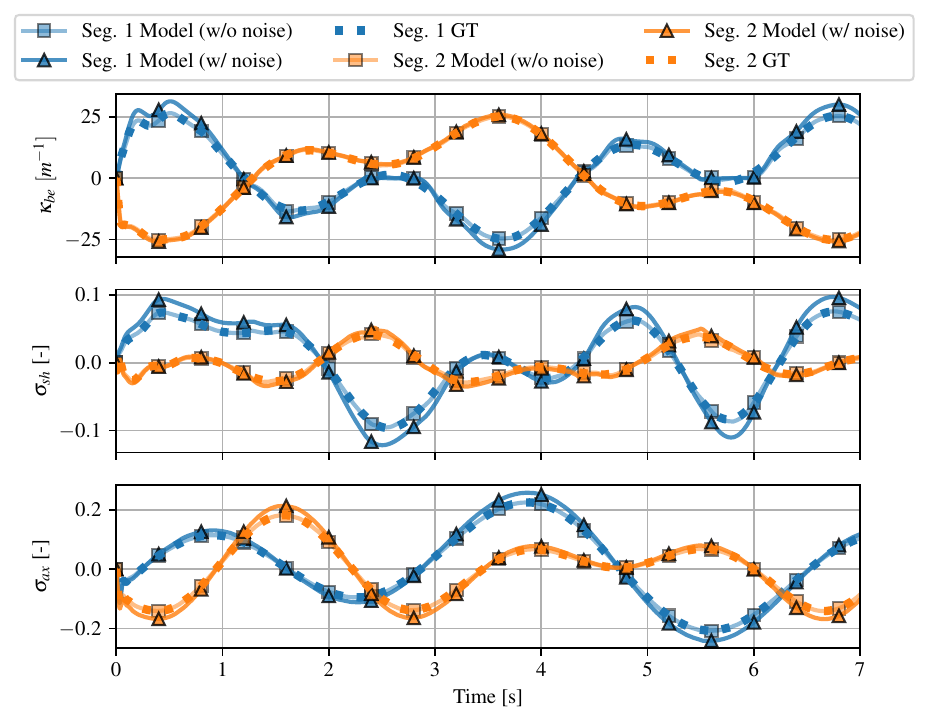}\label{fig:predict_ns-2_subfig-1}
    \vspace{-0.2cm}
    \caption{Verification of the dynamical model with noise for a two-segment PCS soft robot (\emph{Case 2}). The dotted lines denote the ground-truth (GT) trajectory. The blue lines refer to the first segment, while the orange lines are associated with the second segment. 
    }
    \label{fig:predict_ns-2}
    \vspace{-0.2cm}
\end{figure}

\begin{figure*}[ht]
    \centering
    \subfigure[$t=\SI{0.0}{s}$]{\includegraphics[width=0.160\textwidth,trim={5 5 5 5}]{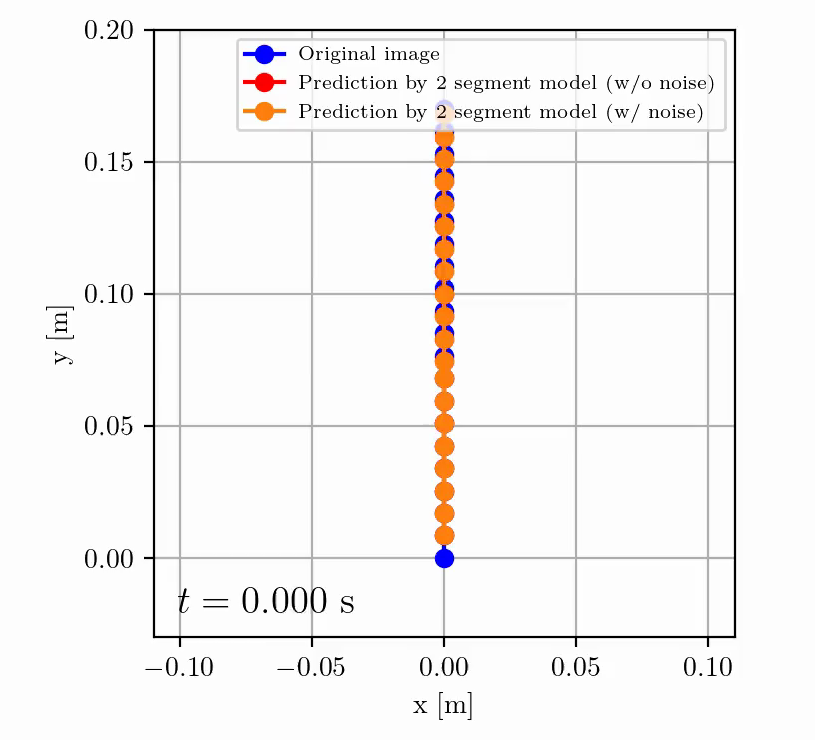}}
    \subfigure[$t=\SI{1.4}{s}$]{\includegraphics[width=0.160\textwidth,trim={5 5 5 5}]{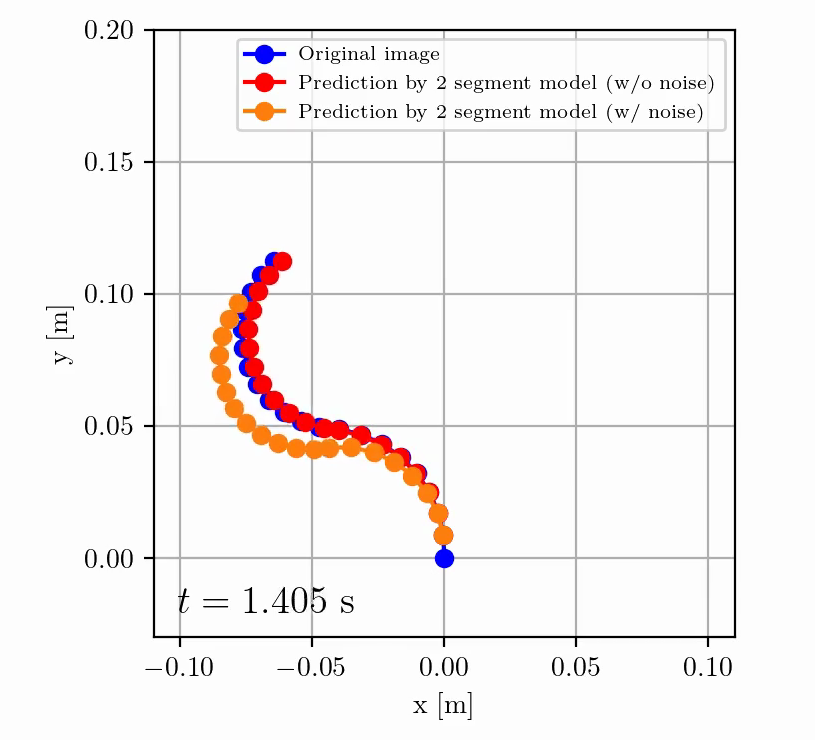}}
    \subfigure[$t=\SI{2.8}{s}$]{\includegraphics[width=0.160\textwidth,trim={5 5 5 5}]{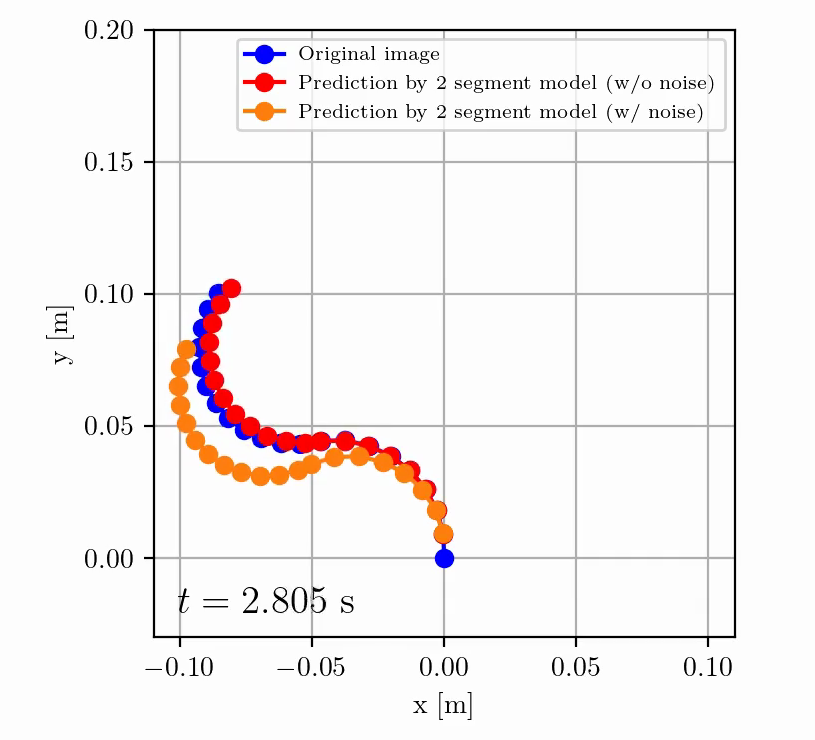}}
    \subfigure[$t=\SI{4.2}{s}$]{\includegraphics[width=0.160\textwidth,trim={5 5 5 5}]{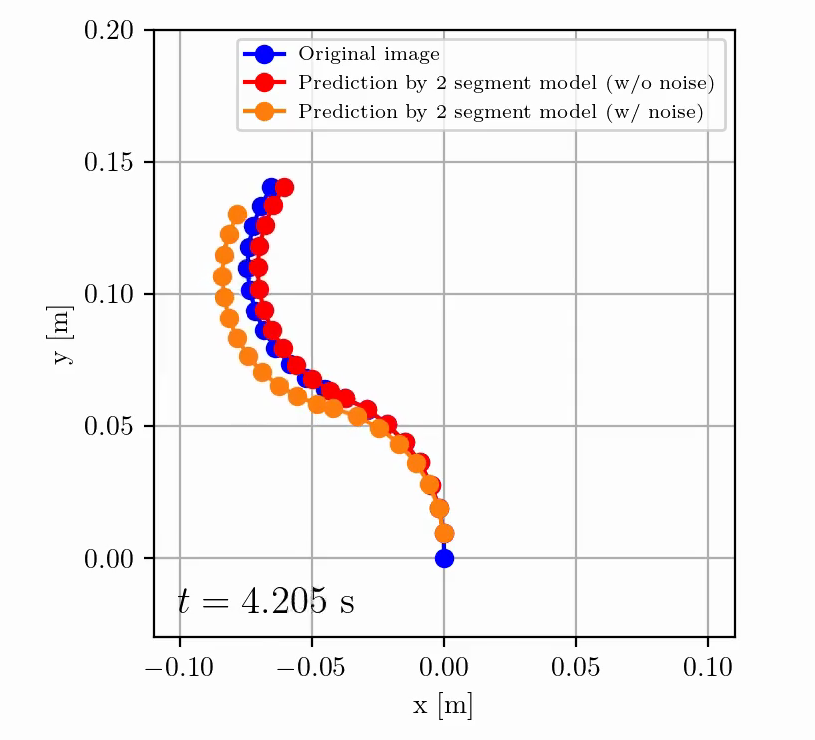}}
    \subfigure[$t=\SI{5.6}{s}$]{\includegraphics[width=0.160\textwidth,trim={5 5 5 5}]{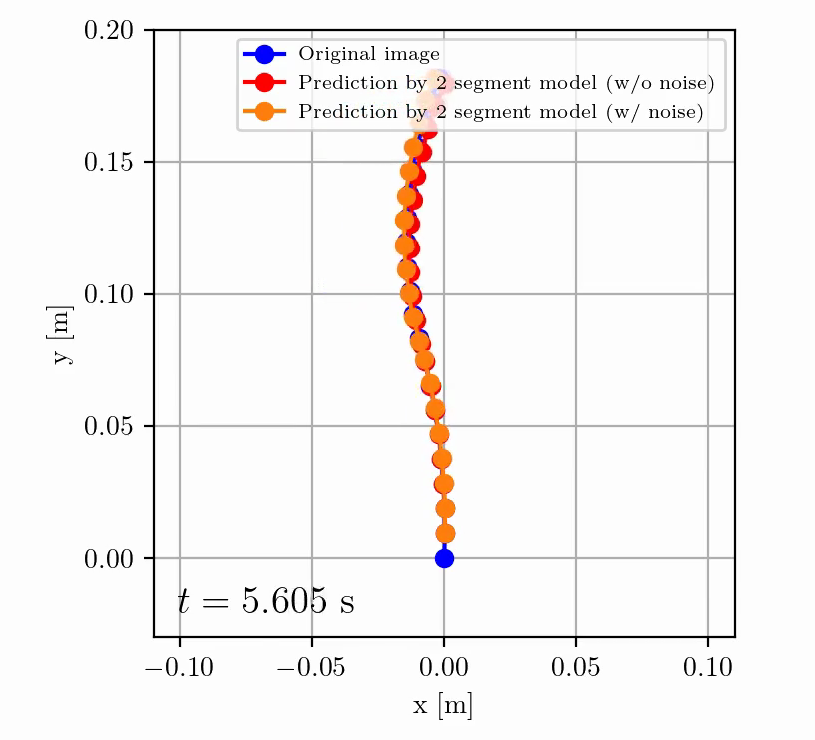}}
    \subfigure[$t=\SI{7.0}{s}$]{\includegraphics[width=0.160\textwidth,trim={5 5 5 5}]{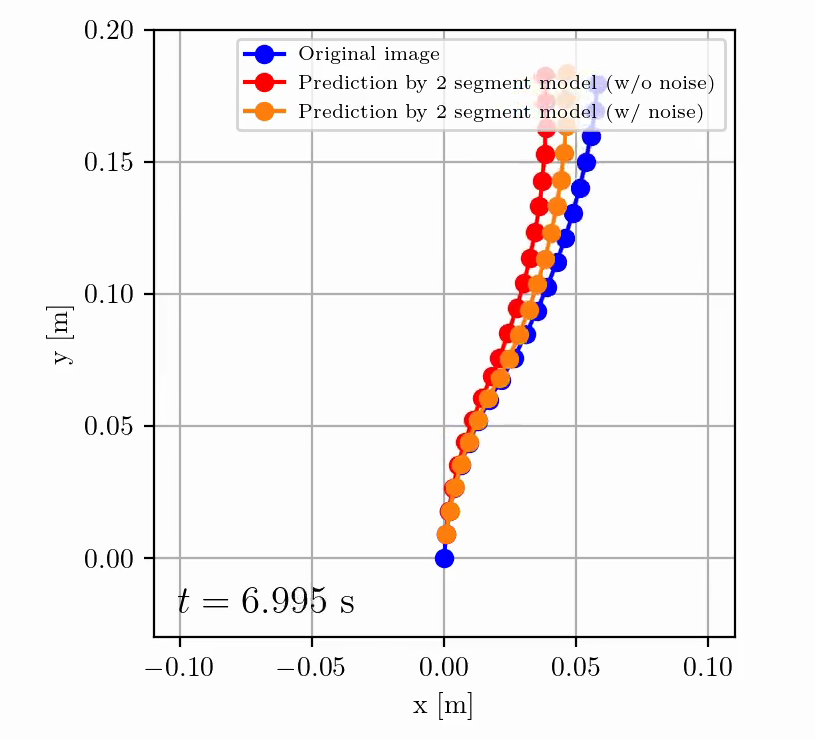}}
    \vspace{-0.2cm}
    \caption{Sequence of stills for the test rollout of the regressed dynamic model trained on the two-segment \ac{PCS} dataset (\emph{Case 2}). The blue and red dots represent the ground truth and estimated shape of the soft robot, respectively. The orange line represents the shape estimated by a model trained on a training set with added measurement noise.}\label{fig:still_seq}
    \vspace{-0.4cm}
\end{figure*}

The results are reported in Tab.~\ref{tab:dyn_results} (top four rows), and the rollout of \emph{Case 2} is included in Fig.~\ref{fig:predict_ns-2}. We also present a sequence of stills of the rollout of \emph{Case 2} in Fig.~\ref{fig:still_seq}.
We conclude that even though the dynamical parameters are regressed on only \SI{4}{s} of robot motion data, the dynamical predictions are extremely accurate on the long horizon of \SI{7}{s} (most control algorithms such as \ac{MPC} operate on a much smaller horizon). The position error for the experiments not involving noise stays below \SI{5}{\percent} in both cases.
When noise is present in the training data, the position error is roughly tripled. Still, we observe that the error is mostly related to the transient terms and model converges during the slower sequences of the trajectory to the ground truth. Most importantly, the learned model, even when trained on noisy data, remains stable, as shown in Fig.~\ref{fig:predict_ns-2}.

\subsubsection{Verification of Strain Sparsification}
\begin{figure}[ht]
    \centering
    \subfigure[Case 4: End-effector pose]{\includegraphics[width=0.49\columnwidth, trim={5 5 5 5}]{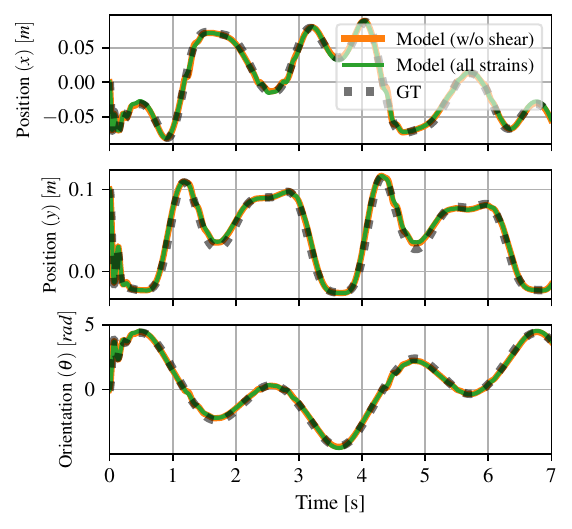}\label{fig:predict_ns-1_high_shear_subfig-2}}
    \subfigure[Case 5: End-effector pose]{\includegraphics[width=0.49\columnwidth, trim={5 5 5 5}]{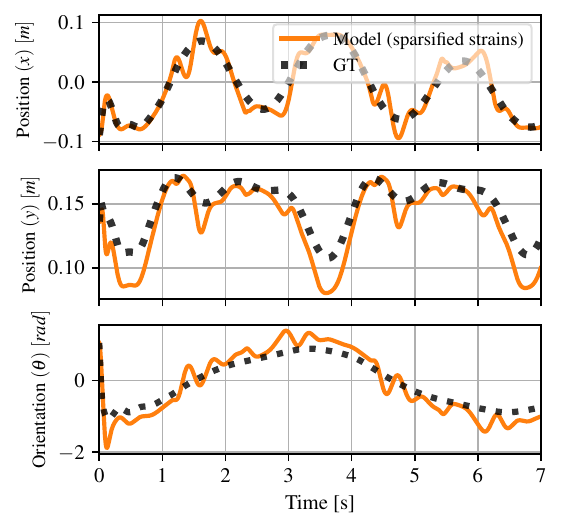}\label{fig:predict_ns-2_high_stiffness_subfig-2}}
    \vspace{-0.2cm}
    \caption{Verification of the strain sparsification algorithm on a one-segment \ac{PCS} robot with shear modulus (\emph{Case 4}) and on a two-segment \ac{PCS} robot where the \nth{1} segment exhibits high axial stiffness and the second segment high shear stiffness. We roll out both the ground truth (dotted line) and the learned (solid line) model dynamics from the same initial condition and for a given sinusoidal actuation sequence.  Verification of the model obtained for Case 4 on a sinusoidal trajectory. The dotted line denotes the ground-truth trajectory, while the green and orange lines correspond to the models trained with and without noise, respectively.
    }
    \label{fig:results:strain_sparsification}
    \vspace{-0.1cm}
\end{figure}

Next, we verify that the strain sparsification algorithm can detect and eliminate strains that do not have a significant effect on the dynamics and can be, therefore, neglected to reduce the model complexity.
For this purpose, we apply the integrated \emph{Dynamic Regression and Strain Sparsification} algorithm to \emph{Cases 4 \& 5}, which exhibit no shear strain and no axial strain (\nth{1}-segment) \& no shear strain (\nth{2}-segment), respectively.
We define the maximum elastic and shear modulus as $E^\mathrm{max} = \SI{100}{MPa}$ which leads to the stiffness thresholds for each segment $K_j^\mathrm{max} = \mathrm{diag}(\SI{12.6}{Nm^2}, \SI{168}{N}, \SI{126}{kN})$.
For example, in \emph{Case 4}, after determining the dynamic parameters during the first iteration, the algorithm detects that the estimated shear stiffness $\hat{K}_\mathrm{sh} = \SI{1200}{N} > K_\mathrm{sh}^\mathrm{max} = \SI{168}{N}$. Therefore, the shear strain is eliminated from the dynamic model, and the dynamic parameters are newly regressed during the next iteration.
Similarly, the algorithm correctly neglects the axial strain for the \nth{1} segment and the shear strain for the \nth{2} segment of \emph{Case 5}.
We visualize the test set rollouts for \emph{Cases 4 \& 5} in Fig.~\ref{fig:results:strain_sparsification} and report the error metrics in the last three rows of Tab.~\ref{tab:dyn_results}.
The results show that in \emph{Case 4}, the model without shear even exhibits a slightly smaller position error than the model that includes all strains. A possible explanation could be that with the Lagrangian being parametrized by fewer basis functions, the coefficients for the remaining strains can be more accurately regressed.
If we compare \emph{Case 2} and \emph{Case 5} (both two-segment \ac{PCS} robots), \emph{Case 5} exhibits significantly increased position and orientation errors. Still, we notice that the model predictions are usable, in particular for shorter horizons.


\begin{table}[htbp]
\centering
\caption{Dynamic regression results: End-effector position and orientation errors $\mathbf{e_\mathrm{p}^{\mathrm{ee}}}, \mathbf{e_{\theta}^{\mathrm{ee}}}$ for the obtained dynamic models evaluated on the \SI{7}{s} sinusoidal test set trajectory. In some cases, we add artificial measurement noise to the training data. For \emph{Case 4}, we present two variants of the learned model: in the first instance, we report the performance of a model that neglects strains as suggested by the dynamic sparsification algorithm. For completeness, we furthermore also state the performance of a model that considers all strains. In \emph{Case 5}, the \nth{1} segment only exhibits bending and shear strains (i.e, no axial strain), and the \nth{2} segment only exhibits bending and axial strains (i.e., no shear strain)}
\label{tab:dyn_results}
\setlength\tabcolsep{2pt}
\begin{scriptsize}
\begin{tabular}{c c c c c }
    \toprule
    \textbf{Case} & \textbf{Meas. Noise} & \textbf{Model Strains}     & $\mathbf{e_\mathrm{p}^{\mathrm{ee}}}$ \textbf{[mm]} & $\mathbf{e_{\theta}^{\mathrm{ee}}}$ \textbf{[rad]} \\
    \midrule
    1: 1S PCS & \xmark                     & All                & $4.89$    & $0.113$             \\ 
    1: 1S PCS & \cmark                        & All                & $13.7$    & $0.307$             \\ 
    \midrule
    2: 2S PCS & \xmark                     & All                & $5.22$    & $0.138$             \\
    2: 2S PCS & \cmark                        & All                & $16.8$    & $0.135$             \\ 
    \midrule
    4: 1S PCS H-SH & \xmark & No shear                     & $4.57$    & $0.099$                                 \\ 
    4: 1S PCS H-SH & \xmark & All                & $5.14$    & $0.116$ \\
    \midrule
    5: 2S PCS H-AX/SH & \xmark & (\cmark, \cmark, \xmark, \cmark, \xmark, \cmark) & $17.9$ & $0.305$\\
    \bottomrule
\end{tabular}
\end{scriptsize}
\end{table}


\begin{figure}[hbt]
    \centering

    \subfigure[Train: End-effector poses]{\includegraphics[width=0.49\columnwidth]{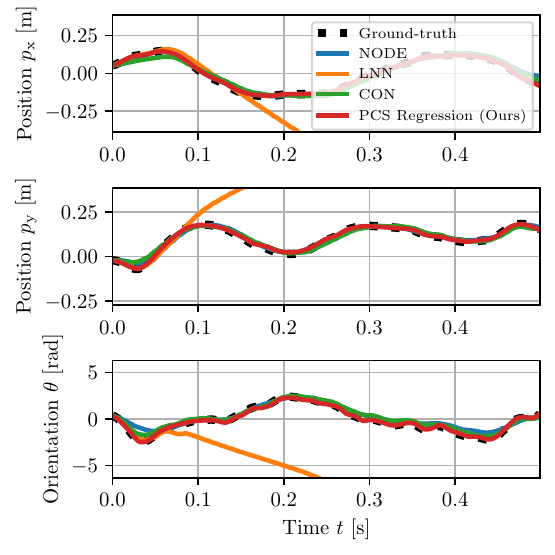}}
    \subfigure[Train: Actuation torques]{\includegraphics[width=0.49\columnwidth]{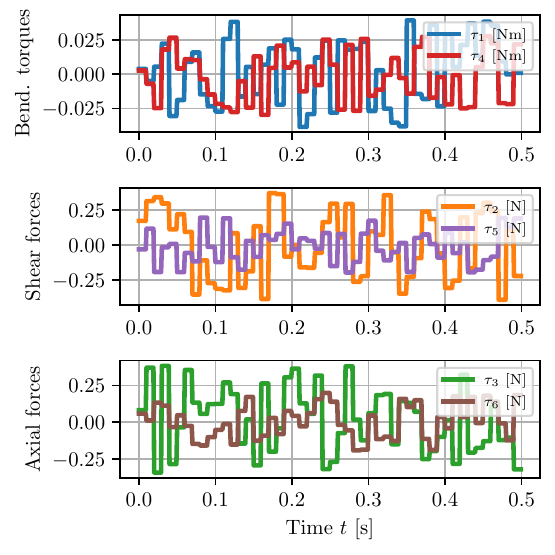}}\\
    \vspace{-0.2cm}
    \subfigure[Test: End-effector poses]{\includegraphics[width=0.49\columnwidth]{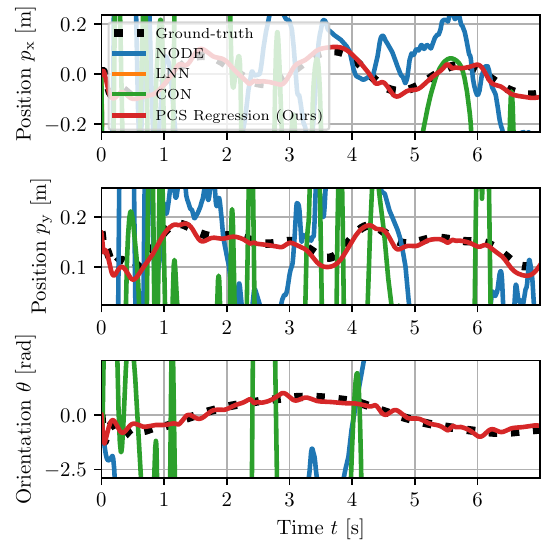}}
    \subfigure[Test: Actuation torques]{\includegraphics[width=0.49\columnwidth]{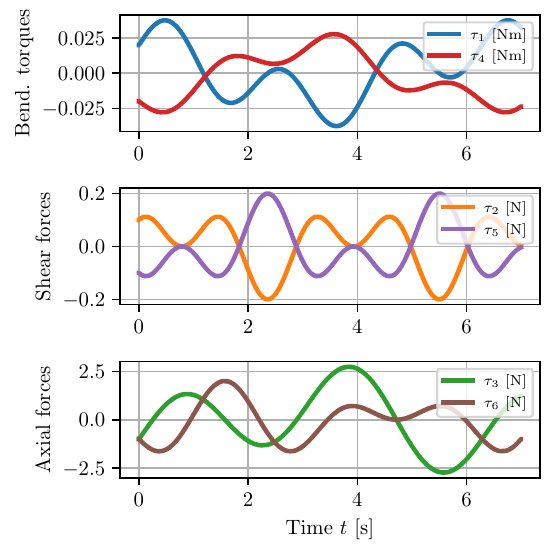}}
    \vspace{-0.2cm}
    \caption{Benchmarking of the proposed method against various machine learning baselines on a PCS soft robot consisting of two segments (\emph{Case 2}): We train the baseline methods on the dynamical evolution of the Cartesian SE(2) poses of three \emph{markers} distributed over the backbone of the soft robot with a total length of \SI{170}{mm}. The upper and lower rows visualize the rollout of all methods on the training and the test set, respectively. The first column shows the evolution end-effector pose. The last column displays the actuation torques that were used to generate the datasets.}
    \label{fig:dynamics_pcs_ns-2_with_baselines}
    \vspace{-0.1cm}
\end{figure}

\begin{table}[htbp]
\centering
\caption{Results for learning dynamics of a two-segment PCS soft robot (Case 2). We report the position ($e_\mathrm{p}^\mathrm{body}$) and orientation ($e_\theta^\mathrm{body}$) metrics that capture the mean shape error averaged over all time steps. The shape error is computed by considering the dynamical evolution of three marker poses with $s_j \in \{ 68, 119, 170 \}$~\si{mm}. We report the metrics for a rollout of the dynamics on both a \SI{0.5}{s} sequence on the training set and the entire (i.e., \SI{7}{s}) test set.}
\label{tab:dynamics_pcs_ns-2_with_baselines}
\setlength\tabcolsep{2pt}
\begin{scriptsize}
\begin{tabular}{c r r r r}
    \toprule
    \textbf{Method} & \textbf{Train} $\mathbf{e_\mathrm{p}^\mathrm{body}}$ & \textbf{Train} $\mathbf{e_\theta^\mathrm{body}}$ & \textbf{Test} $\mathbf{e_\mathrm{p}^\mathrm{body}}$ & \textbf{Test} $\mathbf{e_\theta^\mathrm{body}}$\\
    \midrule
    \ac{NODE} & \SI{10.4}{mm} & \SI{0.27}{rad} & \SI{245.6}{mm} & \SI{12.22}{rad}\\
    \ac{LNN}~\cite{liu2024physics} & \SI{550.8}{mm} & \SI{5.16}{rad} & $\infty$ & $\infty$\\
    \ac{CON}~\cite{stolzle2024input} & \SI{12.9}{mm} & \SI{0.24}{rad} & \SI{789.5}{mm} & \SI{29.49}{rad}\\
    \textbf{PCS Regression (Ours)} & $\mathbf{3.1}$~\si{mm} & $\mathbf{0.04}~\si{rad}$ & $\mathbf{9.8}$~\si{mm} & $\mathbf{0.13}~\si{rad}$\\
    \bottomrule
\end{tabular}
\end{scriptsize}
\vspace{-0.2cm}
\end{table}

\subsection{Benchmarking of Identified Dynamical Model against ML Baselines}
We benchmark the derived dynamical model of \emph{Case 2} (i.e., a two-segment planar \ac{PCS} soft robot) against several models trained using machine learning approaches. Specifically, we consider various learning-based approaches that range from completely data-driven (e.g., \ac{NODE}) over approaches that take into account the structure of Lagrangian systems (e.g., \ac{LNN}, \ac{CON}).
To keep the comparison fair, we define the inputs for all methods as the Cartesian poses $\chi_j \in \mathbb{R}^3$ and the corresponding time derivative $\dot{\chi}_j$ of $N$ discrete \emph{markers} along the backbone. Furthermore, we also provide the actuation torques $\tau \in \mathbb{R}^6$ as a dynamic model input. Therefore, the total model input exhibits a dimensionality of $\mathbb{R}^{6N + 6}$. The task of the dynamic model is to predict the acceleration $\ddot{\chi}(k) \in \mathbb{R}^{3N}$. 
We tried supplying all $21$ markers, which our method also has access to, to the baseline approaches. However, this proved to be infeasible as the problem would become too high-dimensional in terms of the numbers of inputs and outputs, and the baseline approaches would overfit the training set. Therefore, we settled to give the baseline methods access to the pose measurements of $N=3$ markers distributed along the backbone of the robot at $s_j \in \{ 68, 119, 170 \}$~\si{mm}, where $s=\SI{170}{mm}$ corresponds to the end-effector.

\paragraph{Implementation of Baseline Methods}
The \ac{NODE} is parametrized by a six-layer Multilayer Perceptron (MLP) with hidden dimension $256$ and hyperbolic tangent activation function that predicts based on the input $(\chi(t), \dot{\chi}(t), \tau(t))$ the acceleration $\ddot{\chi}$. We remark that with this strategy, we already infuse the prior knowledge that the time derivative of the pose is the velocity, which would not be the case in a naive implementation of a \ac{NODE}.
\acp{CON}~\cite{stolzle2024input} allow for learning of (latent) dynamics of Lagrangian systems with strong stability guarantees (global asymptotic stability / input-to-state stability) by leveraging a network of damped harmonic oscillators that are coupled by a hyperbolic potential. In order to allow for arbitrary placement of the global asymptotically stable equilibrium point, we learn a linear coordinate transformation into the latent coordinates $z = W \, \chi + b \in \mathbb{R}^9$, $\dot{z} = W \dot{\chi}$. 
\acp{LNN} learn the components of the Lagrangian $\mathcal{L}(\chi, \dot{\chi}) = \frac{1}{2} \dot{\chi}^\top M(\chi) \dot{\chi} - \mathcal{U}(\chi)$ such as the mass matrix $M(\chi) \succ 0 \in \mathbb{R}^{9 \times 9}$, the potential energy $\mathcal{U}(\chi) \in \mathbb{R}$, the damping matrix $D \succeq 0 \in \mathbb{R}^{9 \times 9}$ and the actuation matrix $A \in \mathbb{R}^{6 \times 9}$ and subsequently derive the \ac{EOM} as $M(\chi) \ddot{\chi} + \frac{\partial \mathcal{L}}{\partial \chi \partial \dot{\chi}} \dot{\chi} + \frac{\partial \mathcal{U}}{\partial \chi} + D \dot{\chi} = A \tau$ using autodifferentiation.
We regard $A$ and $D$ as trainable weights and parametrize $M(\chi)$ and $\mathcal{U}(\chi)$ with six-layer MLPs with hidden dimension $256$ (softplus activation). We leverage the Cholesky decomposition to make sure that $M(\chi), D \succ 0$.

\paragraph{Training}
The first loss term is a \ac{MSE} between the predicted $\hat{\ddot{\chi}}(k)$ and actual acceleration $\ddot{\chi}(k)$. Additionally, we roll out the trajectories over a horizon of \SI{0.3}{s} and add loss terms that compute the \ac{MSE} error between the predicted states $(\hat{\chi}(k+r), \hat{\dot{\chi}}(k+r))$ and the labels $(\chi(k+r), \dot{\chi}(k+r))$, where $r$ is the index of the rollout step.
As training \acp{LNN} is computationally very demanding due to the need to differentiate w.r.t. both inputs and neural network parameters, we had to reduce the training rollout horizon. 

\paragraph{Results}
We present the benchmarking results in Fig.~\ref{fig:dynamics_pcs_ns-2_with_baselines} and Table~\ref{tab:dynamics_pcs_ns-2_with_baselines}.
We report the performance for rollouts on both the training and the test set.
All methods, except for \ac{LNN}, are able to predict the Cartesian-space evolution of the markers attached to the soft robot backbone decently accurately over the training set trajectory. Still, the our proposed method exhibits an \SI{70}{\percent} to \SI{80}{\percent} lower error than \ac{NODE} and \ac{CON}~\cite{stolzle2024input}. As \ac{LNN} does not exhibit any stability guarantees and it is trained on a relatively short horizon, it diverges from the ground-truth trajectory after roughly \SI{0.15}{s}.
As shown in panels (b) and (d) of Fig.~\ref{fig:dynamics_pcs_ns-2_with_baselines}, the axial actuation forces on the validation set are one order of magnitude higher than in the training set. Therefore, these axial forces can be considered to be out-of-distribution for the trained models. Our proposed method is amazingly able to exhibit very good performance still, while all baseline methods are no longer able to predict the dynamic evolution of the system. \ac{LNN} even becomes fully unstable after a few milliseconds, and we are, therefore, not able to report test errors for this method.

\begin{figure}[ht]
    \centering
    \subfigure[Bending strains]{\includegraphics[width=0.49\columnwidth, trim={5 5 5 5}]{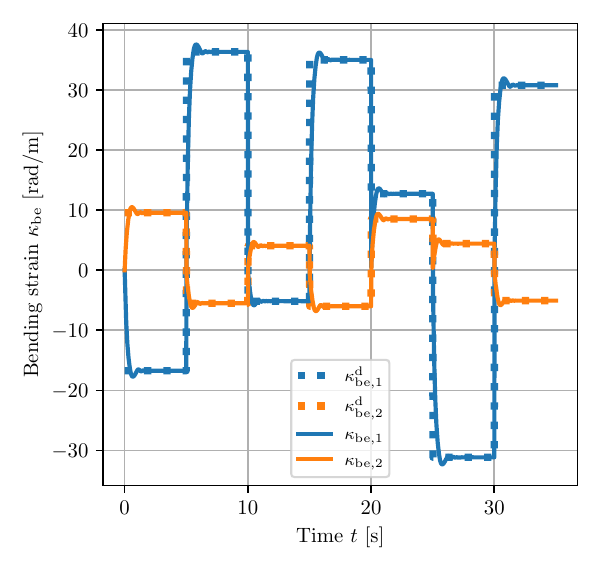}}
    \subfigure[Linear strains]{\includegraphics[width=0.49\columnwidth, trim={5 5 5 5}]{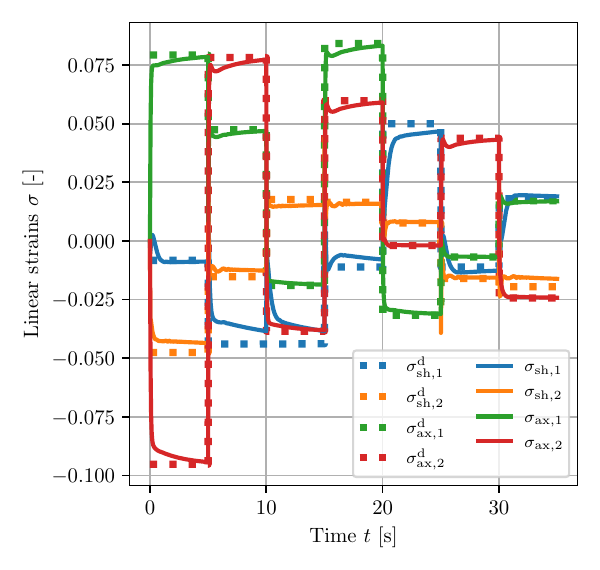}}
    \vspace{-0.2cm}
    \caption{Demo of model-based control of a two-segment soft robot based on the learned dynamical model. We ask the controller to track a sequence of $7$ setpoints $q^\mathrm{d}\in \mathbb{R}^6$ which is denoted with dotted lines. The controller contains a feedforward and feedback term where the feedforward term compensates for the elastic and gravitational forces at the setpoint. 
    }
    \label{fig:results:control:pcs_ns-2}
    \vspace{-0.1cm}
\end{figure}

\subsection{Demo of Model-based Control}
To demonstrate how the derived models can be used in a plug-and-play fashion for model-based control, we simulate the closed-loop dynamics of a simulated two segment \ac{PCS} soft robot (\emph{Case 2}) with configuration $q \in \mathbb{R}^7$ under a P-satI-D+FF~\cite{della2023model, stolzle2023experimental, stolzle2024input} setpoint control policy
\begin{equation}\small
\begin{split}
    \tau(t, q) =& \: \underbrace{\hat{G}(q^\mathrm{d})+\hat{K}q^\mathrm{d}}_{\text{Learned feedforward term}} + \underbrace{K_\mathrm{p} \, (q^\mathrm{d} - q) - K_\mathrm{d} \, \dot{q} + K_\mathrm{i} \, e_\mathrm{int}(t) }_{\text{P-satI-D feedback term~\cite{pustina2022p}}},
\end{split}
\end{equation}
where $K_\mathrm{p}, K_\mathrm{i}, K_\mathrm{d} \in \mathbb{R}^{n \times n}$ are the proportional, integral, and derivative control gains, respectively.
The feedback control term is a PID-like controller with integral saturation~\cite{pustina2022p} and the dimensionless gain $\Upsilon \in \mathbb{R}$ which bounds the integral error at each time step to the interval $(-1, 1)$ and reduces the risk of instability for nonlinear systems
\begin{equation}\small
    e_\mathrm{int}(t, q) = \:  \int_0^t \tanh(\Upsilon (q^\mathrm{d}(t') - q(t'))) \: \mathrm{d}t',
\end{equation}
$\hat{G}(q) \in \mathbb{R}^{6}$ and $\hat{K} \in \mathbb{R}^{6 \times 6}$ are the estimated gravitational forces and stiffness matrix, respectively.
We simulate the closed-loop dynamics with a Tsitouras 5(4) integrator at a timestep of $\SI{0.05}{ms}$.
Please note that we use the ground-truth dynamics as a state transition function.

To verify that the learned model performs well within the model-based control policy, we create a sequence of $7$ randomly sampled setpoints $q^\mathrm{d}(k) \in \mathbb{R}^6$. 
The results in Fig.~\ref{fig:results:control:pcs_ns-2} show that the controller is able to regulate a two-segment planar soft robot effectively. 
The tracking of the bending strain reference is perfect. For the linear strains, we notice small errors in the feedforward term, but the integral control is able to compensate for them and drive the system toward the reference.
We stress that the structure and characteristics of the learned model enabled us to formulate the control policy in closed form easily, and we did not have to resort to techniques such as \ac{MPC} or \ac{RL} as it would be necessary for other model learning techniques (e.g., \acp{LSTM}, \acp{NODE}).
\section{Conclusion}
In this work, we present a data-driven method that utilizes the PCS strain model to derive low-dimensional kinematic and dynamic models for continuum soft robots from discrete backbone pose measurements, outperforming ML-based models like neural networks by maintaining the physical robot structure. This enhancement improves data efficiency and performance beyond the training set, allowing for direct and effective model-based control design. 
Future work will explore expanding this approach to 3D models and real-world applications, aiming to further refine the actuation matrix for underactuated systems.

\bibliographystyle{IEEEtran}
\bibliography{main.bib}

\end{document}